\documentclass[letterpaper]{article} 
\usepackage{aaai25}  
\usepackage{times}  
\usepackage{helvet}  
\usepackage{courier}  
\usepackage[hyphens]{url}  
\usepackage{graphicx} 
\def\UrlFont{\rm}  
\usepackage{natbib}  
\usepackage{multirow} 
\usepackage{caption} 
\usepackage{algorithm}
\usepackage{algorithmic}
\usepackage{svg}
\usepackage{newfloat}
\usepackage{listings}
\DeclareCaptionStyle{ruled}{labelfont=normalfont,labelsep=colon,strut=off} 
\floatstyle{ruled}
\newfloat{listing}{tb}{lst}{}
\floatname{listing}{Listing}
\title{Are VLMs Really Blind (Student Abstract)}
\author{
   Ayush Singh,Mansi Gupta,Shivank Garg }

\affiliations{
    Vision and Language Group, Indian Institute of Technology Roorkee, Roorkee, Uttarakhand, India - 247667 \\
   ayush\_s@mt.iitr.ac.in ,  m\_gupta@ma.iitr.ac.in , shivank\_g@mfs.iitr.ac.in   
}

\begin{document}

\maketitle

\begin{abstract}

Vision Language Models excel in handling a wide range of complex tasks, including Optical Character Recognition (OCR), Visual Question Answering (VQA), and advanced geometric reasoning. However, these models fail to perform well on low-level basic visual tasks which are especially easy for humans. Our goal in this work was to determine if these models are truly ``blind" to geometric reasoning or if there are ways to enhance their capabilities in this area. Our work presents a novel automatic pipeline designed to extract key information from images in response to specific questions. Instead of just relying on direct VQA, we use question-derived keywords to create a caption that highlights important details in the image related to the question. This caption is then used by a language model to provide a precise answer to the question without requiring external fine-tuning. Our code is available at \url{https://github.com/vlgiitr/Are-VLMs-Really-Blind}.

\end{abstract}

\section{Introduction}{
 Large Language Models (LLMs) have revolutionized the field of AI by enabling machines to generate human-like text with remarkable performance. Vision and Language Models (VLMs) extend the capabilities of LLMs to a multimodal setting. These models achieve human-like performance on tasks such as question answering (QA), image captioning, OCR recognition, object detection, and object segmentation.
In our research, we concentrate on the logical reasoning of Vision and Language Models (VLMs) for basic geometry tasks like identifying where circles intersect and counting the number of paths. We break down the Visual Question Answering (VQA) task into two parts: Image Captioning and Text-based Question Answering. Our findings demonstrate that there is an overall improvement in performance when using this approach compared to simple VQA. Additionally, we notice that base models, when incorporated into our approach, outperform models that are specifically fine-tuned for question-answering tasks. We hypothesize that VLMs are not truly blind \cite{rahmanzadehgervi2024visionlanguagemodelsblind} and their performance of simple geometry-related tasks can be improved significantly. We believe captioning followed by QNA works better than direct VQA since the models are largely pre-trained for generating captions and most downstream tasks in general involve the VLM to generate some text. Moreover, the inherent complexity of the Visual Question Answering (VQA) problem arises from its multimodal nature. The challenge lies in combining and interpreting different types of information, such as text and images. This requires the model to understand each modality independently and effectively capture and utilize the interactions between them.
Although the model has shown improvement in performance overall, it consistently struggles with tasks involving counting where we have not been able to achieve a statistically significant improvement in performance. 
This struggle arises largely due to the fact that during training captions that accurately specify the number of objects become extremely rare in the data as the number of objects increases.\cite{paiss2023countclip}

}

\begin{figure*}[!t]
\centering
\resizebox{\textwidth}{!}{%
\begin{tabular}{|c|cc|cc|cc|cc|cc|cc|cc|cc|cc|}
\hline
\multirow{5}{*}{} & \multicolumn{2}{c|}{Task 1} & \multicolumn{2}{c|}{Task 2} & \multicolumn{2}{c|}{Task 3} & \multicolumn{2}{c|}{Task 4} & \multicolumn{2}{c|}{Task 5} & \multicolumn{2}{c|}{Task 6} & \multicolumn{2}{c|}{Task 7} & \multicolumn{2}{c|}{Task 8} & \multicolumn{2}{c|}{Average} \\ \cline{2-19} 
                  & QnA   & QnA+Captions & QnA   & QnA+Captions & QnA   & QnA+Captions & QnA   & QnA+Captions & QnA   & QnA+Captions & QnA   & QnA+Captions & QnA   & QnA+Captions & QnA   & QnA+Captions & QnA & QnA+Captions \\ \hline
Gemini            & 40 & 43        & 78 & 88        & 49 & 51        & 24 & 26        & 33  & 41        & 15 & 14        & 13  & 15         & 65 & 74      & 39.63 & \textbf{44.00} \\ \hline
Paligemma         & 3 & 49        & 90 & 84        & \textbf{60} & 40       & 21 & 34        & 3 & 34        & 0 & \textbf{24}        & 6  & 6         & 37.5 & 55      & 27.56 & \textbf{40.75} \\ \hline
GPT4omini         & 46 & \textbf{71}        & 58 & \textbf{96}        & 18 & 41        & 39 & \textbf{46}        & 40 & \textbf{56}        & 22 & 21        & 48 & \textbf{54}        & \textbf{80}  & 79        & 43.88 & \textbf{58.00} \\ \hline
\end{tabular}%
}
\caption{Results of Various Models When Tested Using Our Pipeline(in percent accuracy)}
\label{tab:tasks}
\end{figure*}

\section{Methodology}{
To evaluate the performance of Vision-Language Models (VLMs) on elementary geometry-related tasks, We selected a set of seven tasks, some of which have been adapted from the Blind dataset\cite{rahmanzadehgervi2024visionlanguagemodelsblind}. These tasks include:
Counting the number of intersections between 2 lines(Task 1),
Checking whether two lines intersect(Task 2),
Counting the number of nested squares in an image(Task 3),
Counting the number of Olympic rings in an image. (Task 4),
Finding the letter circled in red in a word(Task 5),
Counting the number of paths in the subway line task(Task 6), finding the number of rows and columns in a grid(Task 7), and Determining if 2 circles are intersecting(Task 8).
We chose three VLMs—GPT4mini, PaliGemmaBase\footnote{https://huggingface.co/google/paligemma-3b-mix-448} , PaliGemma-QA\footnote{https://huggingface.co/BUAADreamer/PaliGemma-3B-Chat-v0.2}, and Geminiflash—to test our hypothesis. The tasks were divided into two groups: counting-related tasks (tasks 1,3,4,5 and 7) and non-counting geometry tasks (tasks 2, 5, and 8). This categorization allowed us to systematically analyze the model's performance across different types of geometric reasoning.
In our experiments, we took a two-step approach. First, we generated image captions by inputting task-specific keywords derived from the Llama 3.1 Instruction fine-tuned model\footnote{https://huggingface.co/meta-llama/Meta-Llama-3.1-8B-Instruct}. These keywords were extracted by prompting the model to produce concise, 1-2 word summaries that encapsulate the essence of each question. For instance, for the task of counting intersections of circles, we used the prompt: ``Provide a detailed caption for the image using keywords: count, intersections, circles." This method aimed to guide the captioning process toward the specific question we planned to ask later.
After generating the captions, we fed them back into an LLM(Geminiflash) and asked the corresponding questions for each task. We then compared the performance of this pipeline to the traditional approach, where the image is directly input into the model, and the question is posed without prior captioning.
This methodology allowed us to assess the effectiveness of using keyword-directed image captioning as a precursor to question answering, and to evaluate whether this approach improves VLM's performance on both counting and non-counting geometry-related tasks.

\begin{figure}[h]
\centering
\includegraphics[width=0.95\columnwidth, height=0.12\textheight]{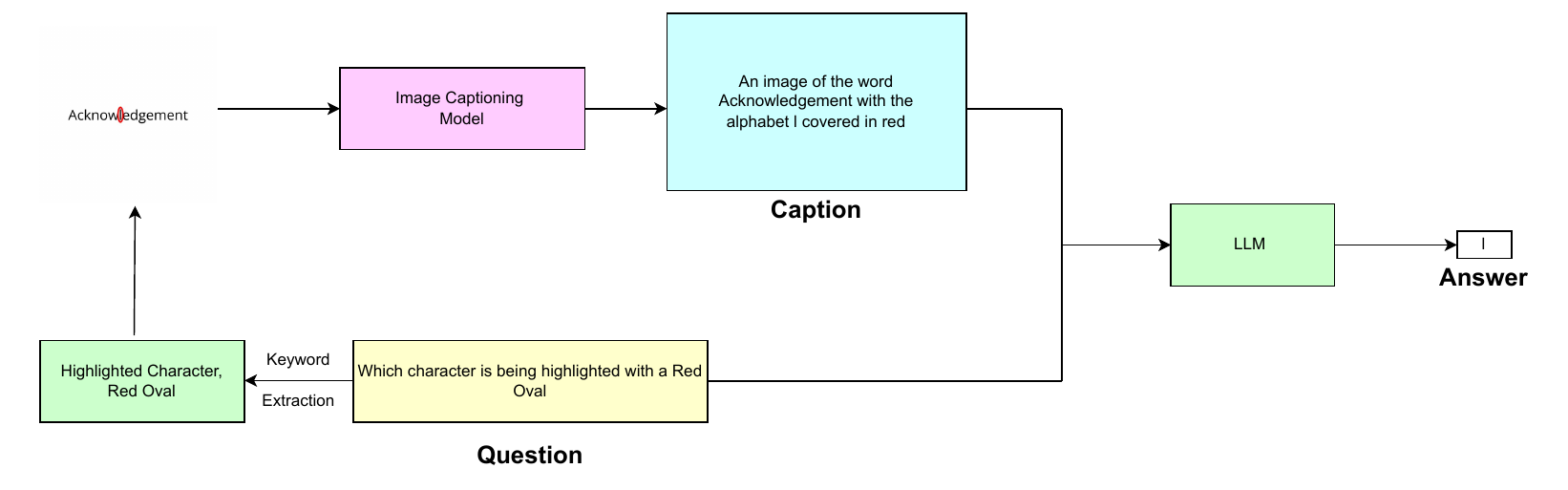} 
\caption{Our approach consists of three key steps. First, we extract relevant keywords from the question using an LLM. These keywords are then provided to a VLM to generate a descriptive caption. Finally, we utilize another LLM to interpret the generated caption and formulate an answer to the original question.}
\label{approach}
\end{figure}
}

\section{Results} {
Figure \ref{tab:tasks} presents the accuracy metrics for each model under two different settings:  direct question answering without the intermediate captioning step, and keyword-driven captioning followed by question answering. Our results demonstrate that the accuracy obtained through the keyword captioning pipeline consistently meets or surpasses that obtained during direct VQA.
Notably, the average accuracy across all tasks is higher when the keyword-driven captioning pipeline is employed. This supports our first hypothesis that guiding the caption generation process with relevant keywords can enhance the model's overall performance in VQA tasks. Furthermore, our second hypothesis is substantiated by the data: the improvement in accuracy is particularly significant and uniform for non-counting geometry-related tasks, whereas the average increase in accuracy for counting-related tasks is less visible and somewhat random and not consistent.

}

\section{Future Scope} {
There is significant potential for further research to strengthen and validate our hypothesis. Testing our approach on more advanced, large-scale VLMs, given sufficient computational resources, would help evaluate its scalability and robustness across diverse architectures. Additionally, using specialized datasets like MathVision\cite{wang2024measuring} could provide a more rigorous assessment, particularly for geometry-related tasks. Exploring alternative methods, such as advanced prompt engineering or incorporating domain-specific knowledge, may improve caption accuracy and relevance. Moreover, to enhance explainability, replacing the QA model with interpretable alternatives, like graph-based models, could be investigated.
}

\section{Conclusion} {
Our research provides a thorough assessment of zero-shot models for image captioning in the context of Visual Question Answering (VQA). Our analysis shows that incorporating question-driven image captions into the VQA process significantly improves overall performance. We found a notable enhancement in performance for geometry-related tasks when using keyword-based captioning. This finding suggests that VLMs are not entirely ``blind" to images; rather, they can be guided to perform better on geometry-related tasks through carefully designed pipelines. However, our results also highlight the ongoing challenge VLMs face in tasks related to counting. These tasks remain particularly challenging and random for VLMs. This difficulty arises from the models' inherent limitations. Consequently, despite the enhancements in the visual aspect of the task,  VLMs continue to fall short in accurately performing counting tasks.
}

\bibliography{aaai25.bib} 
\end{document}